\documentclass[conference]{IEEEtran}
\IEEEoverridecommandlockouts
\usepackage{cite}
\usepackage{amsmath,amssymb,amsfonts}
\usepackage{algorithmic}
\usepackage{graphicx}
\usepackage{textcomp}
\usepackage{xcolor}
\usepackage{stfloats}
\usepackage{subcaption}
\usepackage{float}
\def\BibTeX{{\rm B\kern-.05em{\sc i\kern-.025em b}\kern-.08em
    T\kern-.1667em\lower.7ex\hbox{E}\kern-.125emX}}

\makeatletter

\def\@seccntformat#1{\csname the#1\endcsname.\quad}

\makeatother

\begin{document}

\title{Transformer-Based Multi-Step Bilateral Knee Angle Prediction and
Gait Phase Recognition Using Wearable IMU Sensors}

\author{
\IEEEauthorblockN{
R. Anitha\IEEEauthorrefmark{1},
Vismaya S\IEEEauthorrefmark{2}
Vinoth S\IEEEauthorrefmark{1},
Vikash Kumar\IEEEauthorrefmark{1},
K. Mohanavelu\IEEEauthorrefmark{1}
}

\IEEEauthorblockA{\IEEEauthorrefmark{1}
DEBEL, DRDO, Bangalore, India\\
}

\IEEEauthorblockA{\IEEEauthorrefmark{2}
Department of CSE, JSS Academy of Technical Education,
Bangalore, India\\
}
}

\maketitle

\begin{abstract}

Anticipating future knee joint movements is important for
rehabilitation, wearable robotics, gait analysis, and assistive
systems, where motion prediction can help compensate for
sensing and control delays. This study presents a
Transformer-based multi-step sequence-to-sequence framework
for predicting bilateral knee flexion--extension angles using
wearable IMU data acquired from the Xsens MVN Awinda system.
Motion data from 25 healthy participants performing
level-ground walking, inclined and declined walking, stair
ascent, and stair descent were used for evaluation.
The model processes 128 historical time steps to predict
60 future knee-angle samples. A rule-based pseudo-labeling
method is applied to the predicted knee-angle trajectories
to provide gait-phase interpretation without manually
annotated gait labels. The original experimental evaluation
yielded a mean absolute error (MAE) of 1.95$^\circ$ and
a root mean square error (RMSE) of 2.65$^\circ$.
A separate five-fold subject-level cross-validation
evaluation yielded an MAE of 17.54 $\pm$ 6.75$^\circ$
and an RMSE of 21.79 $\pm$ 8.23$^\circ$, respectively.
The gait-phase interpretation achieved pseudo-label
agreement accuracies of 84.41\% and 83.22\% for the
right and left knees, respectively. The results demonstrate
the feasibility of combining multi-step bilateral knee-angle
prediction with rule-based gait-phase interpretation.
The framework provides a basis for further investigation
of predictive gait analysis and wearable assistive systems.

\end{abstract}


\section{Introduction}

The prediction of human joint movement has become increasingly important in domains such as rehabilitation, wearable robotics, gait monitoring, and assistive technology \cite{ref9,ref10,ref11}. In numerous applications, a delay often exists between the detection of a person's movement and the generation of an appropriate response. Anticipating future joint movements can reduce this delay by allowing the system to foresee upcoming movements and deliver timely assistance. This capability is particularly advantageous in rehabilitation, where a precise estimation of future joint motion can enhance patient support, movement monitoring, and the control of wearable assistive devices. 

Among the lower limb joints, the knee plays a vital role in maintaining balance and facilitating forward movement during walking \cite{ref10,ref11}.The continuous variation in knee flexion and extension throughout the gait cycle provides valuable insights into an individual's walking pattern. Consequently, predicting future knee joint angles can improve gait analysis, advance rehabilitation efforts, and aid in the development of intelligent wearable devices. 

Several studies have explored machine and deep learning techniques for gait analysis and lower-limb motion prediction. Despite these promising advances, several challenges remain to be addressed. Most existing studies focus on estimating the current or immediate future joint angle rather than conducting long-horizon multi-step predictions\cite{ref22,ref25}. Moreover, gait-phase recognition is frequently treated as an independent task, necessitating manually annotated gait labels, which increases the annotation effort and limits scalability\cite{ref21}. Additionally, recurrent architectures may have difficulty in effectively capturing the long-range temporal dependencies inherent in complex gait sequences. These limitations diminish the applicability of existing approaches for continuous gait monitoring, rehabilitation, and predictive control in wearable assistance systems.

\begin{table*}[!t]
\caption{Comparison of the Proposed Framework with Recent Studies on Wearable Sensor-Based Lower-Limb Motion Prediction}
\label{tab:literature_comparison}
\centering
\footnotesize
\renewcommand{\arraystretch}{1.25}
\resizebox{\textwidth}{!}{
\begin{tabular}{|p{2.3cm}|p{1.9cm}|p{2.6cm}|p{2.0cm}|c|c|c|c|p{3.6cm}|}
\hline
\textbf{Reference} &
\textbf{Model} &
\textbf{Dataset} &
\textbf{Prediction Target} &
\textbf{Bilateral} &
\textbf{Multi-Step} &
\textbf{Gait Phase} &
\textbf{Manual Labels} &
\textbf{Main Contribution} \\
\hline

Wang et al. (2023) &
TCN--BiLSTM &
19 healthy subjects (IMU) &
Joint angle \& joint moment estimation &
No &
No &
No &
Yes &
Estimated lower-limb joint angles and joint moments from wearable IMU signals using a hybrid temporal deep learning model. \\
\hline

Mohammadi Moghadam et al. (2023) &
CNN, RF, SVM, MARS &
17 healthy subjects (IMU + EMG) &
Kinematics, kinetics, muscle force &
No &
No &
No &
Yes &
Compared conventional machine learning models for predicting lower-limb biomechanics from wearable sensors. \\
\hline

Ding et al. (2024) &
Self-Attention + LSTM &
10 healthy subjects (IMU) &
Knee and ankle joint-angle estimation &
No &
No &
No &
Yes &
Introduced a self-attention network for continuous lower-limb joint-angle estimation during different locomotion modes. \\
\hline

Daryakenari and Farizeh (2025) &
Transformer Neural Network &
22 healthy subjects (sEMG + IMU) &
Lower-limb joint-angle and moment prediction &
Yes &
Yes &
No &
Yes &
Predicted lower-limb kinematics and kinetics over short- and long-term horizons using wearable sensors. \\
\hline


\textbf{Proposed Work} &
\textbf{Transformer Seq2Seq} &
\textbf{25 healthy subjects (Xsens MVN Awinda\textsuperscript{\textregistered} IMU)} &
\textbf{Future bilateral knee-angle prediction} &
\textbf{Yes} &
\textbf{Yes} &
\textbf{Yes} &
\textbf{No} &
\textbf{A unified Transformer-based framework for long-horizon bilateral knee-angle prediction and automatic gait-phase recognition using rule-based pseudo-labeling without manual annotation.} \\
\hline

\end{tabular}
}
\end{table*}

Table~\ref{tab:literature_comparison} summarizes representative studies on the prediction of lower-limb motion using wearable sensors. Current methodologies predominantly emphasize the estimation of joint angles in real time or the prediction over short time horizons, often treating gait-phase recognition as an independent task that necessitates manually annotated gait labels. Although recent transformer-based approaches have shown enhanced capabilities in temporal modeling, they fail to concurrently offer long-horizon bilateral knee-angle prediction and automatic gait-phase interpretation within a single framework. Conversely, the proposed framework integrates multistep bilateral knee-angle prediction with rule-based gait-phase recognition, thereby obviating the requirement for manual gait annotation and enhancing both the interpretability and practical applicability of motion prediction in rehabilitation and wearable assistive systems.

This study introduces a Transformer-based sequence-to-sequence framework aimed at predicting future bilateral knee joint angles. This prediction utilizes wearable IMU data obtained from the Xsens MVN Awinda\textsuperscript{\textregistered} motion capture system \cite{ref7}. The proposed model effectively captures long-range temporal dependencies between historical lower-limb motion and subsequent bilateral knee joint behavior.

\subsection{Novel Contributions}
The principal contributions of this study are outlined as follows:
\begin{enumerate}
\item A Transformer-based sequence-to-sequence architecture has been developed for the long-horizon multi-step prediction of bilateral knee flexion-extension angles utilizing wearable IMU sensor data.
\item A unified framework is proposed that concurrently performs future knee-angle prediction and gait-phase recognition, thereby offering both quantitative motion forecasting and interpretable gait analysis.
\item A rule-based pseudo-labeling strategy is introduced to automatically generate gait-phase labels from predicted knee-angle trajectories, thus obviating the need for manual gait annotation.
\item The proposed framework undergoes comprehensive evaluation using regression and gait-phase classification metrics and is assessed in comparison with an LSTM-based baseline to demonstrate its efficacy for future wearable rehabilitation and assistive applications.
\end{enumerate}

\section{Experimental Protocol}

\subsection{Participants}

Motion data were collected from 25 healthy adult male and female
participants. Five locomotion activities were
performed: level-ground walking, inclined walking, declined walking,
stair ascent, and stair descent. For level-ground treadmill walking,
the speed was maintained at 5~km/h, while the remaining activities
were performed at a normal walking pace without a recorded
specific speed. Participant-level demographic information, including
age, height, and body mass, was not available in the dataset used for
the present analysis; therefore, demographic statistics could not be
reported.

\begin{figure*}[!t]
\centering
\includegraphics[width=0.80\textwidth]{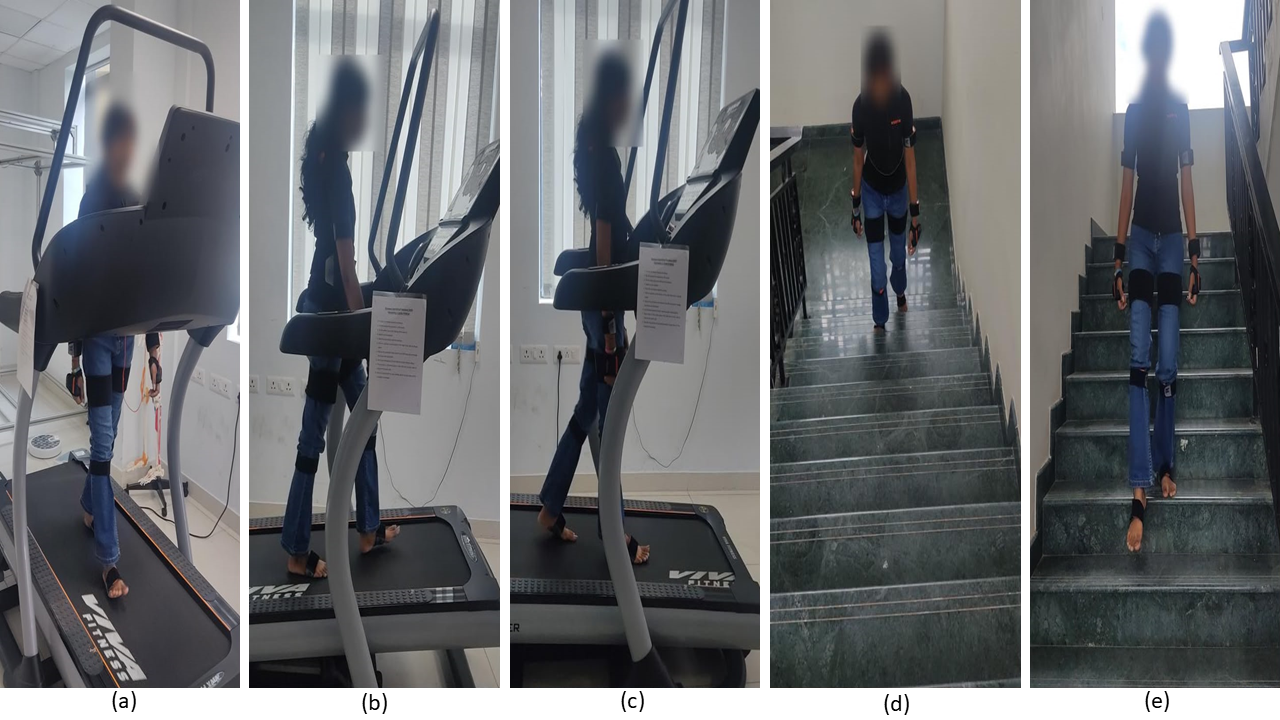}
\caption{Representative gait activities performed during data collection:
(a) level-ground walking, (b) inclined walking, (c) declined walking,
(d) stair ascent, and (e) stair descent while wearing the Xsens MVN
Awinda\textsuperscript{\textregistered} IMU system.}
\label{fig:experimental_activities}
\end{figure*}

Figure~\ref{fig:experimental_activities} illustrates the locomotion
activities performed during data collection.

\subsection{Data Acquisition System}

Motion data were acquired using the Xsens MVN Awinda\textsuperscript{\textregistered}
system \cite{ref7}, which consists of 17 wireless inertial measurement
units (IMUs) positioned at standardized anatomical locations. The
system provides measurements of acceleration, angular velocity,
orientation, and joint kinematics. Data were recorded at a sampling
rate of 100 Hz throughout all experimental trials.

\subsection{Sensor Calibration and Data Recording}

Prior to data collection, the Xsens\textsuperscript{\textregistered} system was calibrated using the standard N-pose procedure described in the Xsens\textsuperscript{\textregistered} documentation \cite{ref7,ref8}, where participants stood upright with their arms positioned slightly away from the sides of the body. A short walking trial was then performed to verify sensor alignment and data quality. Re-calibration was carried out whenever sensor drift or misalignment was observed during recording.

\begin{figure}[htbp]
\centerline{\includegraphics[width=\linewidth]{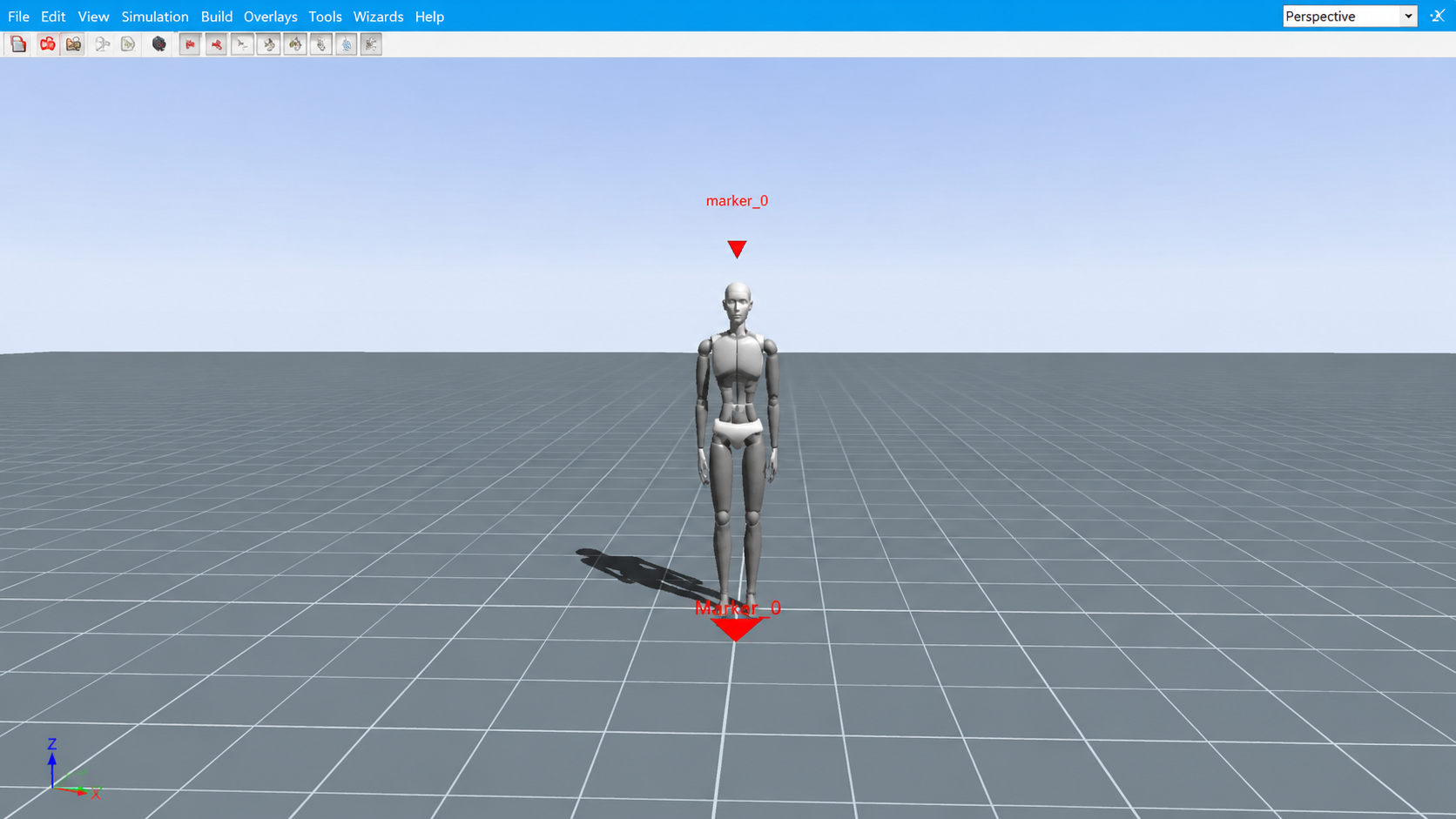}}
\caption{N-pose calibration performed before gait data acquisition}
\label{fig:n_pose_calibration}
\end{figure}

\subsection{Sensor Placement}

\begin{figure}[htbp]
\centering
\includegraphics[width=\linewidth]{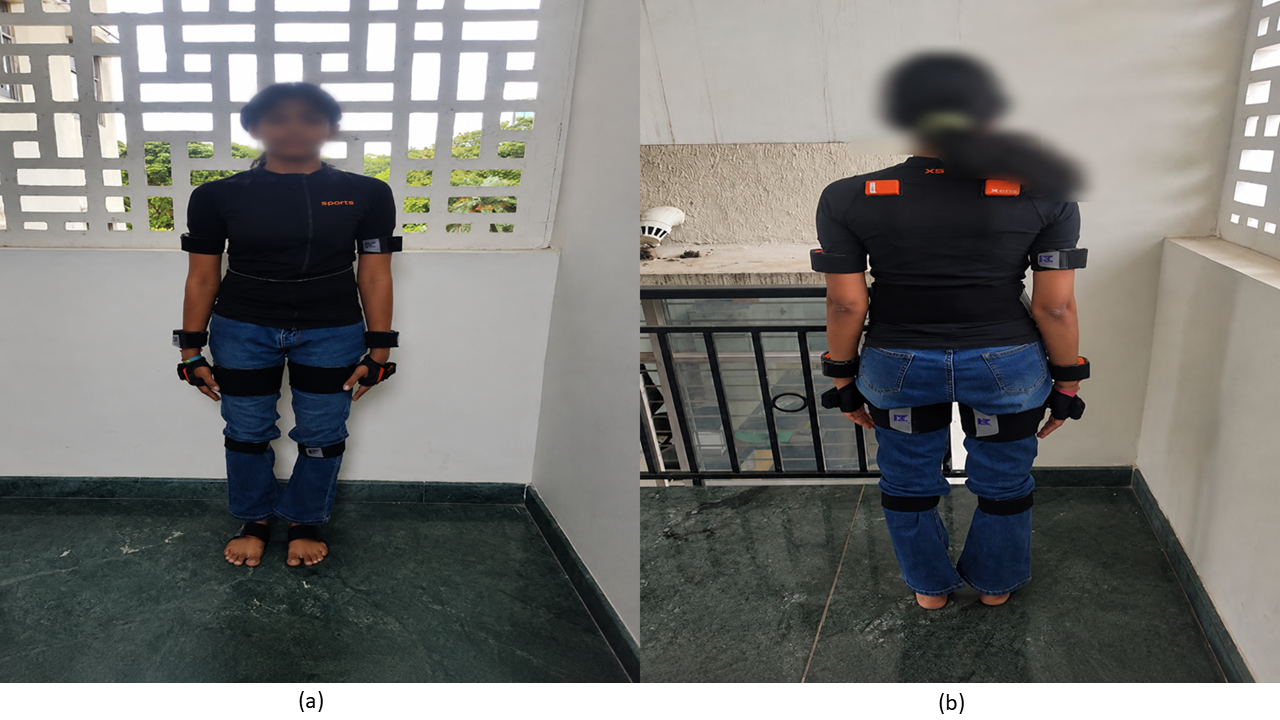}
\caption{Wearable IMU sensor arrangement for gait data collection: (a) front view and (b) back view}
\label{fig:imu_sensor_placement}
\end{figure}

The IMU sensors were positioned at predefined anatomical locations
including the head, sternum, pelvis, upper arms, forearms, hands,
thighs, shanks, and feet. Proper placement ensured consistent
measurement of body-segment motion and lower-limb joint kinematics.
Figure~\ref{fig:imu_sensor_placement} illustrates the sensor
configuration used during the experiments.

\subsection{Data Processing Environment}

The recorded motion data were processed using MVN Analyze Pro software
\cite{ref8}. Body-segment orientation, angular velocity, free
acceleration, and bilateral knee flexion--extension angles provided by
the Xsens system were exported for subsequent preprocessing and model
development. Although surface EMG signals were recorded simultaneously,
only IMU-derived measurements were used as inputs to the proposed
Transformer-based prediction framework.
\begin{figure}[!t]
\centering
\includegraphics[width=1.00\columnwidth]{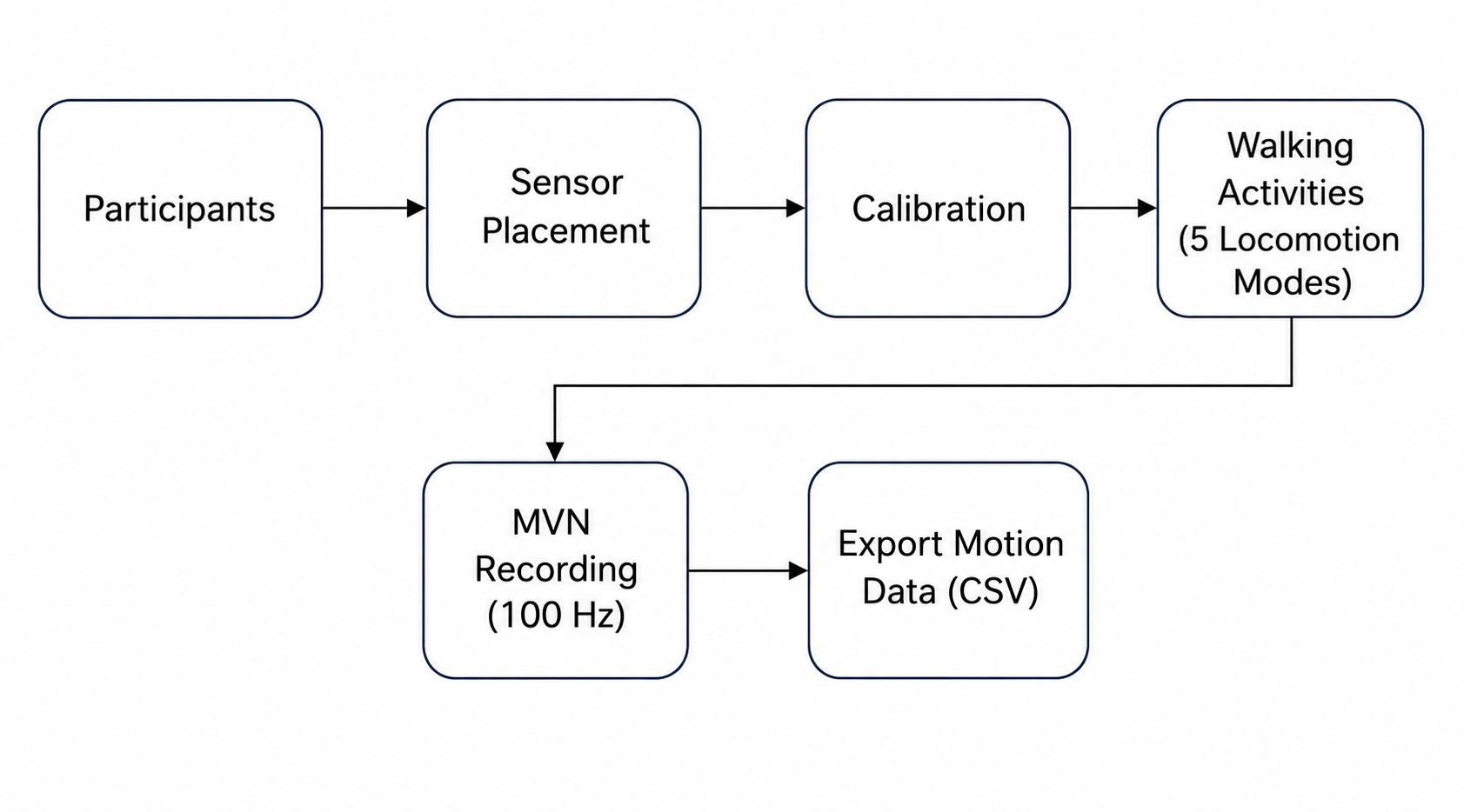}
\caption{Data acquisition pipeline using the Xsens MVN Awinda\textsuperscript{\textregistered} IMU system.}
\label{fig:data_acquisition_pipeline}
\end{figure}

\section{Proposed Methodology}

This section describes the overall framework used for knee-angle prediction and gait-phase interpretation.

\subsection{Overview}

The proposed framework integrates future bilateral knee-angle prediction and automatic gait-phase interpretation within a unified Transformer-based architecture. In the first stage, preprocessed IMU-derived features are organized into temporal sequences and provided to a Transformer-based sequence-to-sequence model to predict future left and right knee flexion--extension angles. The model learns movement patterns from previous lower-limb motion. It then uses this information to estimate future knee joint angles. In the second stage, the predicted knee-angle sequences are analyzed using a rule-based pseudo-labeling procedure described in Section~\ref{subsec:gait_phase_interpretation} to derive gait phases.

\begin{figure*}[!t]
\centering
\includegraphics[
    width=0.70\textwidth]{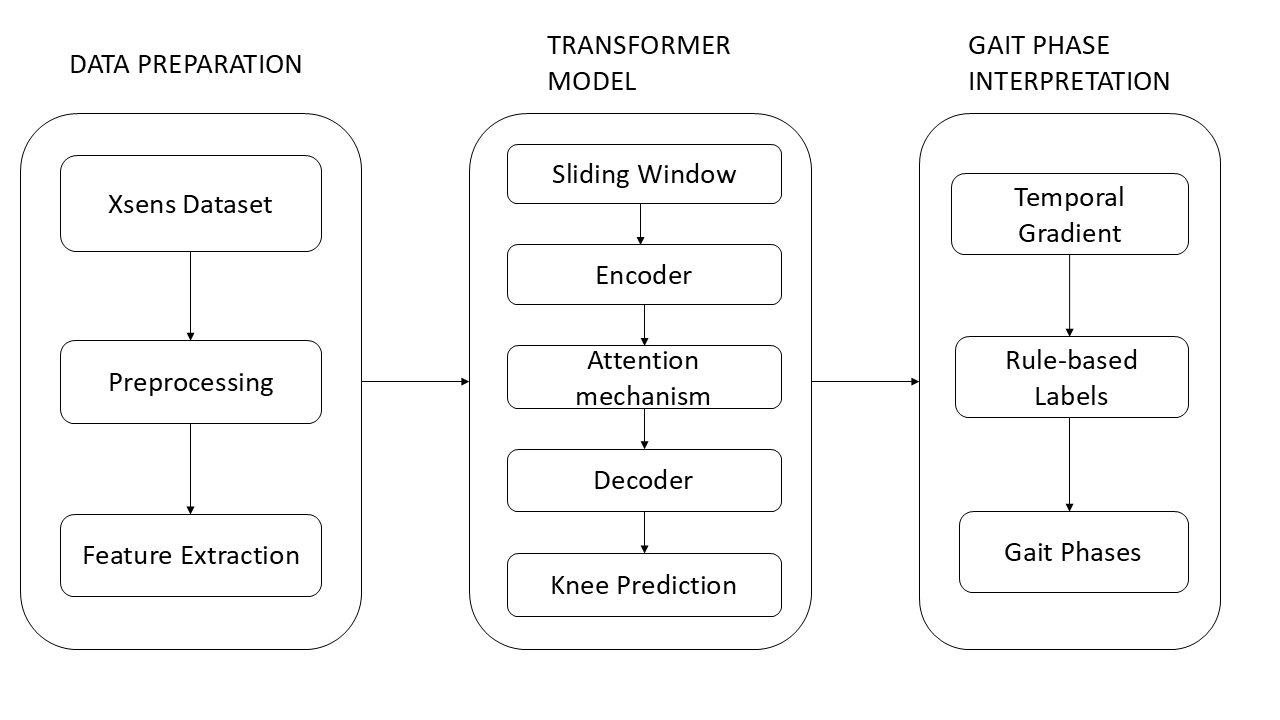}
\caption{Workflow of the proposed knee joint angle prediction framework}
\label{fig:methodology_workflow}
\end{figure*}

Figure~\ref{fig:methodology_workflow} illustrates the overall
workflow, from IMU data preprocessing and sliding-window generation
to Transformer-based prediction and gait-phase interpretation.

\subsection{Input Features and Prediction Targets}

The prediction model uses motion data obtained from wearable IMU sensors positioned on the lower limbs. The selected features include free acceleration, angular velocity, and segment orientation measurements collected from the thigh, shank, and foot segments of both legs. These signals describe lower-limb movement dynamics. They are commonly used in gait analysis to characterize joint motion and segment behaviour during walking\cite{ref10,ref11}.

The recorded sensor measurements were synchronized and organized into multivariate time-series sequences before being provided to the prediction model. Since the data were collected from both lower limbs, the input features capture bilateral movement patterns. They also capture the coordination between different body segments during gait.

The prediction targets consist of the knee flexion–extension angles of both the left and right knees. Table~\ref{tab:input_features} summarizes the input signals, body segments, prediction targets, and data representation used by the proposed Transformer-based prediction framework. These joint angles were obtained from the kinematic outputs generated by the Xsens MVN Awinda\textsuperscript{\textregistered} system and expressed in degrees. By predicting knee motion for both limbs simultaneously, the model is able to learn the temporal relationships between past movement patterns and future joint behaviour during different locomotion activities.

\begin{table}[!t]
\caption{Summary of Input Features and Prediction Targets}
\label{tab:input_features}
\centering
\footnotesize
\renewcommand{\arraystretch}{1.2}
\begin{tabular}{|p{2.8cm}|p{4.2cm}|}
\hline
\textbf{Component} & \textbf{Description} \\
\hline

Sensor Platform &
Xsens MVN Awinda\textsuperscript{\textregistered} IMU System \\

\hline

Sampling Frequency &
100 Hz \\

\hline

Input Signals &
Free acceleration, angular velocity, and segment orientation \\

\hline

Body Segments &
Left and right thigh, shank, and foot \\

\hline

Input Representation &
Multivariate time-series sequence \\

\hline

Prediction Target &
Left and right knee flexion--extension angles (degrees) \\

\hline

Prediction Type &
Multi-step sequence-to-sequence forecasting \\

\hline

Output Horizon &
60 future samples (0.6 s) \\

\hline
\end{tabular}
\end{table}

\subsection{Data Preprocessing}

As illustrated in Fig.~\ref{fig:methodology_workflow}, after data
preprocessing, the continuous IMU measurements were converted into
supervised learning samples using a sliding-window strategy. The
training, validation, and test windows were generated independently
within their respective partitions to prevent windows from crossing
partition boundaries.

The raw IMU signals contained high-frequency fluctuations caused by sensor noise and measurement variability. To suppress these unwanted components while preserving the underlying gait characteristics, Gaussian smoothing was applied to the sensor signals. This filtering approach reduced high-frequency noise while maintaining the temporal structure of the movement data, thereby improving the quality of the input sequences used for prediction.

Following signal smoothing, the selected input features were standardized
using z-score normalization to ensure a common numerical scale. To
prevent information leakage between the training and evaluation
partitions, the mean and standard deviation for each feature were
computed exclusively from the training partition of each evaluation
fold. These training-derived parameters were then applied unchanged to
the corresponding training, validation, and held-out test partitions.
For each input feature, the standardized value was calculated as

\begin{equation}
z=\frac{x-\mu_{\mathrm{train}}}{\sigma_{\mathrm{train}}}
\end{equation}

where $x$ denotes the original feature value, and
$\mu_{\mathrm{train}}$ and $\sigma_{\mathrm{train}}$ denote the mean
and standard deviation calculated exclusively from the training
partition, respectively. Features with a standard deviation below
$10^{-8}$ were assigned a standard deviation of 1 to avoid numerical
instability.

The bilateral knee-angle target variables were standardized using the
same training-only procedure. The target normalization parameters were
calculated from the training targets of each fold and subsequently
applied to the validation and test targets. Before calculating the
final regression metrics, the predicted knee angles were transformed
back to their original degree scale using the corresponding
training-derived parameters.

After standardization, the processed IMU measurements were organized into synchronized multivariate time-series sequences while preserving their temporal order. These processed sequences served as the input to the subsequent sliding-window sequence generation stage shown in Fig.~\ref{fig:methodology_workflow}. The preprocessing pipeline reduced the influence of measurement noise, standardized the feature space, and preserved the temporal dependencies required for future knee-angle prediction.

\begin{figure}[!t]
    \centering
    \includegraphics[width=1.00\columnwidth]{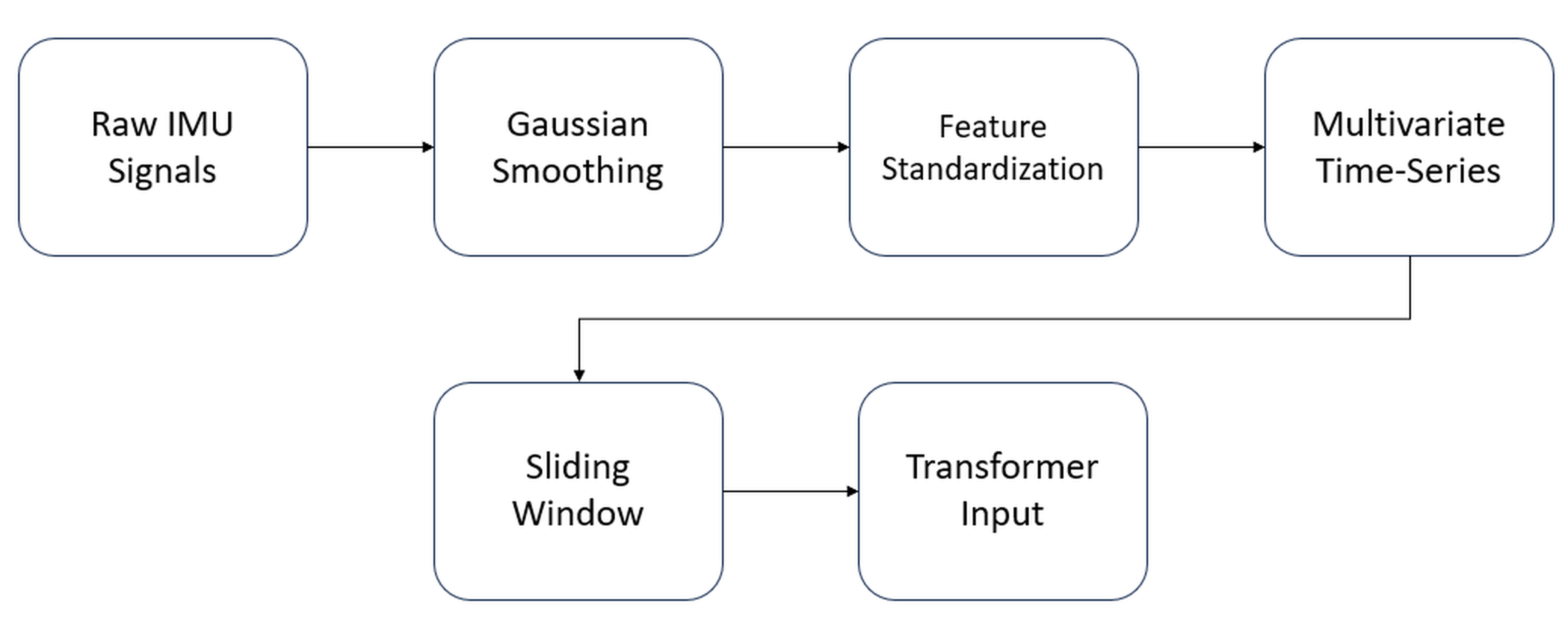}
    \caption{Data preprocessing pipeline for Transformer model input generation.}
    \label{fig:data_preprocessing_pipeline}
\end{figure}

\subsection{Subject-Level Cross-Validation and Data Partitioning}

To prevent information leakage between training and evaluation data, the 
dataset was partitioned at the subject level prior to sliding-window 
generation. A grouped five-fold cross-validation strategy was 
adopted, where the 25 participants were divided into five groups of five 
subjects. In each fold, four groups (20 subjects) were used for model 
development, while the remaining group (5 subjects) was held out 
exclusively for testing. This procedure was repeated across all five 
folds so that every subject served as a test subject exactly once.

For each fold, the training subjects were further divided temporally, 
with the first 80\% of each recording used for training and the final 
20\% reserved for validation. Sliding windows were generated 
independently within the training, validation, and test partitions to 
prevent windows from crossing partition boundaries. Feature and target 
normalization parameters were calculated only from the training data and 
then applied unchanged to the validation and test data. This protocol 
ensures that the reported performance reflects generalization to 
previously unseen subjects.

\subsection{Sliding Window Sequence Generation}

As illustrated in Fig.~\ref{fig:methodology_workflow}, after data preprocessing, the IMU measurements were converted into supervised learning samples using a sliding-window strategy. Since gait signals are continuous time-series data, they must be segmented into fixed-length input-output sequences before being provided to the Transformer-based sequence-to-sequence model. This approach preserves the temporal continuity of human movement while generating a large number of overlapping training samples.

In this study, each input sequence consisted of 128 consecutive time steps, corresponding to approximately 1.28~s of historical gait information recorded at a sampling frequency of 100~Hz. The prediction target comprised the subsequent 60 time steps, representing approximately 0.6~s of future bilateral knee motion. To account for sensing and processing delays encountered in practical wearable systems, a forward shift of two time steps (20~ms) was introduced between the input sequence and the prediction horizon.

The generated input and target sequences are defined as

\begin{equation}
X_t=[x_t,x_{t+1},\ldots,x_{t+127}]
\end{equation}

\begin{equation}
Y_t=[y_{t+130},y_{t+131},\ldots,y_{t+189}]
\end{equation}

where $X_t$ represents the 128-step multivariate IMU input sequence
and $Y_t$ represents the 60-step future bilateral knee-angle
sequence. The two-step shift provides a 20~ms prediction gap between
the observation window and the forecast horizon.

\begin{figure*}[!t]
\centering \includegraphics[width=0.70\linewidth]{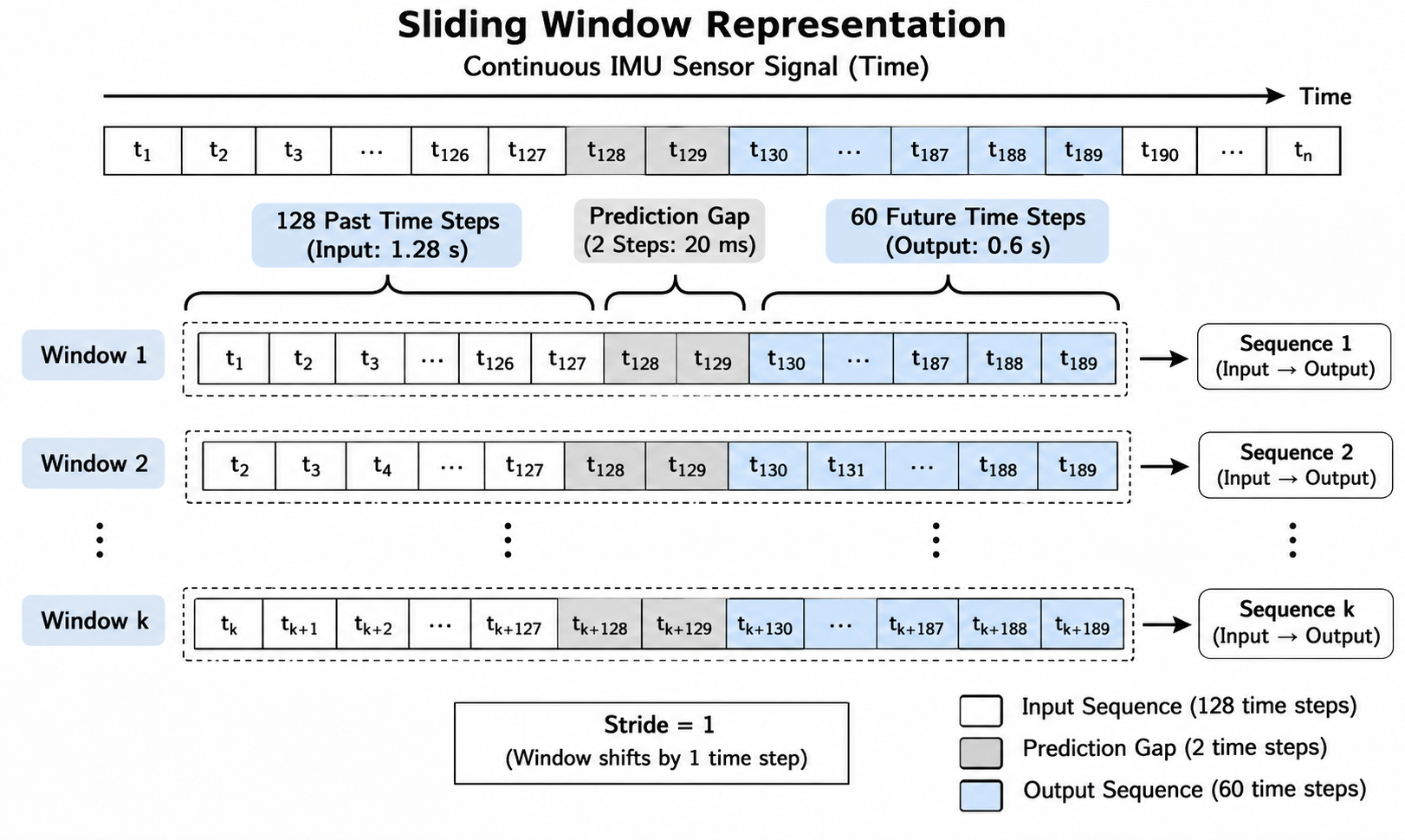}
\caption{Sliding window representation for generating input–output sequences used in sequence-to-sequence knee joint angle prediction}
\label{fig:sliding_window_rep}
\end{figure*}

Figure~\ref{fig:sliding_window_rep} illustrates the sliding-window
sequence generation. Within each training, validation, and test
partition, the window was moved sequentially with a fixed stride to
generate overlapping samples while preserving the temporal
relationships in the gait signals. Sliding windows were generated
independently within each partition, ensuring that no window crossed
a training, validation, or test boundary.

\subsection{Transformer-Based Knee Angle Prediction}

A Transformer-based sequence-to-sequence model was used to predict future knee flexion–extension angles from the processed IMU measurements. The transformer architecture was selected because it can effectively model temporal relationships within sequential gait data using attention mechanisms \cite{ref1,ref2}.

The input sequences generated using the sliding-window approach were first transformed into embedded representations and combined with positional information to preserve the temporal order. These embedded sequences were then passed through the encoder, which learned meaningful temporal patterns from past lower-limb movements. An attention mechanism was employed to identify the most relevant time steps and movement characteristics that contribute to future knee motion.

The attention mechanism used in the transformer architecture can be expressed as

\begin{equation}
Attention(Q,K,V)=
\text{softmax}
\left(
\frac{QK^{T}}
{\sqrt{d_k}}
\right)V
\end{equation}

where $Q$, $K$, and $V$ represent the query, key, and value matrices, respectively, and $d_k$ denotes the dimensionality of the key vectors \cite{ref1}.

The encoded representations were subsequently provided to the decoder, which generated the future knee-angle sequence for both the left and right knees. By learning the relationship between previous movements and future joint behaviour, the model was able to estimate upcoming knee flexion–extension angles over the prediction horizon.

Model training was performed using supervised learning, where the predicted knee-angle sequences were compared with the reference knee-angle values obtained from the motion capture system. The network parameters were optimized using the Adam optimizer \cite{ref6} to minimize the prediction error and improve future motion estimates over successive training iterations.
\\

\subsubsection{Mathematical Formulation of the Proposed Transformer-Based Model}

The proposed Transformer-based framework treats future knee-angle prediction as a sequence-to-sequence learning problem. The objective is to learn the relationship between previously observed IMU measurements and future bilateral knee joint angles.

Let the IMU measurements collected at time step $t$ be represented as

\begin{equation}
x_t=[x_t^1,x_t^2,\ldots,x_t^F]
\end{equation}

where $F$ denotes the total number of sensor features extracted from the lower-limb IMUs.

Using the sliding-window approach, 128 consecutive timesteps are grouped together to form an input sequence

\begin{equation}
X=[x_1,x_2,\ldots,x_{128}]
\end{equation}

such that

\begin{equation}
X \in \mathbb{R}^{128 \times F}
\end{equation}

This input sequence represents 1.28 seconds of past gait information recorded at a sampling frequency of 100 Hz.

The prediction target consists of future bilateral knee-angle values over the next 60 timesteps

\begin{equation}
Y=[y_1,y_2,\ldots,y_{60}]
\end{equation}

where

\begin{equation}
y_t=[\theta_R(t),\theta_L(t)]
\end{equation}

and

\begin{equation}
Y \in \mathbb{R}^{60\times2}
\end{equation}

with $\theta_R$ and $\theta_L$ representing the right and left knee flexion-extension angles respectively.

\vspace{2mm}
\textbf{A: Input Projection}

The raw IMU feature matrix is first projected into a higher-dimensional latent representation using a learnable linear transformation

\begin{equation}
E=XW_E+b_E
\end{equation}

where

\begin{equation}
W_E \in \mathbb{R}^{F\times96}
\end{equation}

The value 96 corresponds to the model dimension used in the proposed architecture.

The projected representation becomes

\begin{equation}
E \in \mathbb{R}^{128\times96}
\end{equation}

This operation converts the raw sensor features into a representation that can be processed by the Transformer model.

\vspace{2mm}
\textbf{B: Positional Encoding}

Since the transformer architecture does not inherently preserve temporal order, positional information is added to the embedded sequence.

\begin{equation}
Z=E+PE
\end{equation}

where $PE$ denotes the sinusoidal positional encoding matrix.

The positional encoding is computed as

\begin{equation}
PE(pos,2i)
=
\sin
\left(
\frac{pos}
{10000^{\frac{2i}{d_{model}}}}
\right)
\end{equation}

\begin{equation}
PE(pos,2i+1)
=
\cos
\left(
\frac{pos}
{10000^{\frac{2i}{d_{model}}}}
\right)
\end{equation}

As a result, the model can distinguish between earlier and later gait events within the sequence.

\vspace{2mm}
\textbf{C: Query, Key and Value Generation}

The position-aware representation is transformed into Query, Key and Value matrices

\begin{equation}
Q=ZW_Q
\end{equation}

\begin{equation}
K=ZW_K
\end{equation}

\begin{equation}
V=ZW_V
\end{equation}

where $W_Q$, $W_K$ and $W_V$ are trainable projection matrices.

The resulting matrices satisfy

\begin{equation}
Q,K,V \in \mathbb{R}^{128\times96}
\end{equation}

\vspace{2mm}
\textbf{D: Similarity Matrix Computation}

To determine the relationship between different timesteps, the Query matrix is multiplied by the transpose of the Key matrix

\begin{equation}
S=QK^T
\end{equation}

where

\begin{equation}
S \in \mathbb{R}^{128\times128}
\end{equation}

Each element of $S$ represents the similarity between two timesteps in the gait sequence.

\begin{equation}
S_{ij}=q_i \cdot k_j
\end{equation}

Larger values indicate stronger temporal relationships.

\vspace{2mm}
\textbf{E: Scaled Dot-Product Attention}

The similarity scores are scaled to improve numerical stability

\begin{equation}
S'
=
\frac{QK^T}
{\sqrt{d_k}}
\end{equation}

The proposed model uses four attention heads and a model dimension of 96.

Therefore

\begin{equation}
d_k=\frac{96}{4}=24
\end{equation}

\vspace{2mm}
\textbf{F: Attention Weight Computation}

The scaled similarity scores are converted into attention weights using the Softmax function

\begin{equation}
A=
\text{softmax}
\left(
\frac{QK^T}
{\sqrt{d_k}}
\right)
\end{equation}

where

\begin{equation}
A\in\mathbb{R}^{128\times128}
\end{equation}

Each row of $A$ represents the relative importance of all timesteps when processing a particular timestep.

\vspace{2mm}
\textbf{G: Context Vector Generation}

The attention weights are multiplied with the Value matrix to obtain contextual representations

\begin{equation}
H=
AV
\end{equation}

or equivalently

\begin{equation}
H=
\text{softmax}
\left(
\frac{QK^T}
{\sqrt{d_k}}
\right)V
\end{equation}

This enables the model to focus on the most informative gait events when learning movement patterns.

\vspace{2mm}
\textbf{H: Multi-Head Attention}

Instead of learning a single attention representation, the model employs four independent attention heads

\begin{equation}
head_i=
Attention(Q_i,K_i,V_i)
\end{equation}

for

\begin{equation}
i=1,2,3,4
\end{equation}

The outputs are concatenated and projected

\begin{equation}
M=
Concat(head_1,head_2,head_3,head_4)W_O
\end{equation}

Using multiple attention heads enables the model to learn different temporal dependencies within the same gait sequence.

\vspace{2mm}
\textbf{I: Encoder Block Processing}

The attention output is combined with the original input using a residual connection followed by layer normalization

\begin{equation}
M_1=
LayerNorm(M+Z)
\end{equation}

The normalized representation is then passed through the feed-forward network

\begin{equation}
F_1=
ReLU(M_1W_1+b_1)
\end{equation}

where

\begin{equation}
W_1\in\mathbb{R}^{96\times192}
\end{equation}

The hidden representation becomes

\begin{equation}
F_1\in\mathbb{R}^{128\times192}
\end{equation}

A second linear transformation is applied

\begin{equation}
F=
F_1W_2+b_2
\end{equation}

where

\begin{equation}
W_2\in\mathbb{R}^{192\times96}
\end{equation}

The output is again combined with a residual connection and normalized

\begin{equation}
E_{out}
=
LayerNorm(F+M_1)
\end{equation}

This completes one encoder layer.

Since the proposed architecture contains three encoder layers, the same process is repeated three times.

\vspace{2mm}
\textbf{J: Encoder Memory Representation}

After passing through all encoder layers, the final memory representation becomes

\begin{equation}
H_{enc}
=
Encoder(X)
\end{equation}

where

\begin{equation}
H_{enc}
\in
\mathbb{R}^{128\times96}
\end{equation}

This encoder output contains the temporal information learned from the previous 128 time steps and serves as memory for the decoder.

\vspace{2mm}
\textbf{K: Decoder Processing}

A learnable decoder query matrix is initialized

\begin{equation}
D_0
\in
\mathbb{R}^{60\times96}
\end{equation}

which corresponds to the 60 future time steps that need to be predicted.

Positional information is added

\begin{equation}
D_1=D_0+PE_d
\end{equation}

The decoder first performs masked self-attention

\begin{equation}
D_2=
Attention(D_1,D_1,D_1)
\end{equation}

followed by

\begin{equation}
D_3=
LayerNorm(D_2+D_1)
\end{equation}

The decoder then performs cross-attention using the encoder memory

\begin{equation}
D_4=
Attention(D_3,H_{enc},H_{enc})
\end{equation}

followed by

\begin{equation}
D_5=
LayerNorm(D_4+D_3)
\end{equation}

The decoder representation then passes through the feed-forward network

\begin{equation}
D_6=
ReLU(D_5W_1+b_1)
\end{equation}

\begin{equation}
D_7=
D_6W_2+b_2
\end{equation}

Finally,

\begin{equation}
D=
LayerNorm(D_7+D_5)
\end{equation}

The same process is performed across both decoder layers to generate future knee joint representations.

\vspace{2mm}
\textbf{L: Output Projection and Knee-Angle Prediction}

The final decoder representation is projected into future bilateral knee-angle predictions

\begin{equation}
\hat{Y}
=
DW_O+b_O
\end{equation}

where

\begin{equation}
W_O
\in
\mathbb{R}^{96\times2}
\end{equation}

The predicted output becomes

\begin{equation}
\hat{Y}
\in
\mathbb{R}^{60\times2}
\end{equation}

Each row of the output matrix contains the predicted right and left knee flexion-extension angles for a particular future timestep.

The complete model therefore learns the mapping

\begin{equation}
f:
\mathbb{R}^{128\times F}
\rightarrow
\mathbb{R}^{60\times2}
\end{equation}

or equivalently

\begin{equation}
\hat{Y}=f(X)
\end{equation}

During training, the Mean Absolute Error (MAE) loss function is minimized

\begin{equation}
L=
\frac{1}{120}
\sum_{t=1}^{60}
\left(
|\theta_R^t-\hat{\theta}_R^t|
+
|\theta_L^t-\hat{\theta}_L^t|
\right)
\end{equation}

During training, the Adam optimizer is used to update the model parameters iteratively until the prediction error becomes minimum.

The Transformer model configuration used in the experiments is
summarized in Table~\ref{tab:hyperparameters}. These hyperparameters were selected based on preliminary experiments to achieve a balance between prediction accuracy, training stability, and computational efficiency for multi-step bilateral knee-angle prediction.

\begin{table}[!t]
\caption{Transformer Model Hyperparameters}
\label{tab:hyperparameters}
\centering
\footnotesize
\renewcommand{\arraystretch}{1.2}
\begin{tabular}{lc}
\hline
\textbf{Parameter} & \textbf{Value} \\
\hline
Input Sequence Length & 128 \\
Prediction Horizon & 60 \\
Embedding Dimension & 96 \\
Encoder Layers & 3 \\
Decoder Layers & 2 \\
Attention Heads & 4 \\
Feed-Forward Dimension & 192 \\
Optimizer & Adam \\
Loss Function & MAE \\
Learning Rate & 0.0005 \\
Batch Size & 64 \\
Epochs & 20 \\
Dropout & 0.1 \\
Evaluation Protocol & Five-Fold Subject-Level Cross-Validation \\
Test Subjects per Fold & 5 \\
Validation Split & 20\% temporal tail of training subjects \\
\hline
\end{tabular}
\end{table}

\begin{figure*}[!t]
\centering
\includegraphics[width=0.70\textwidth]{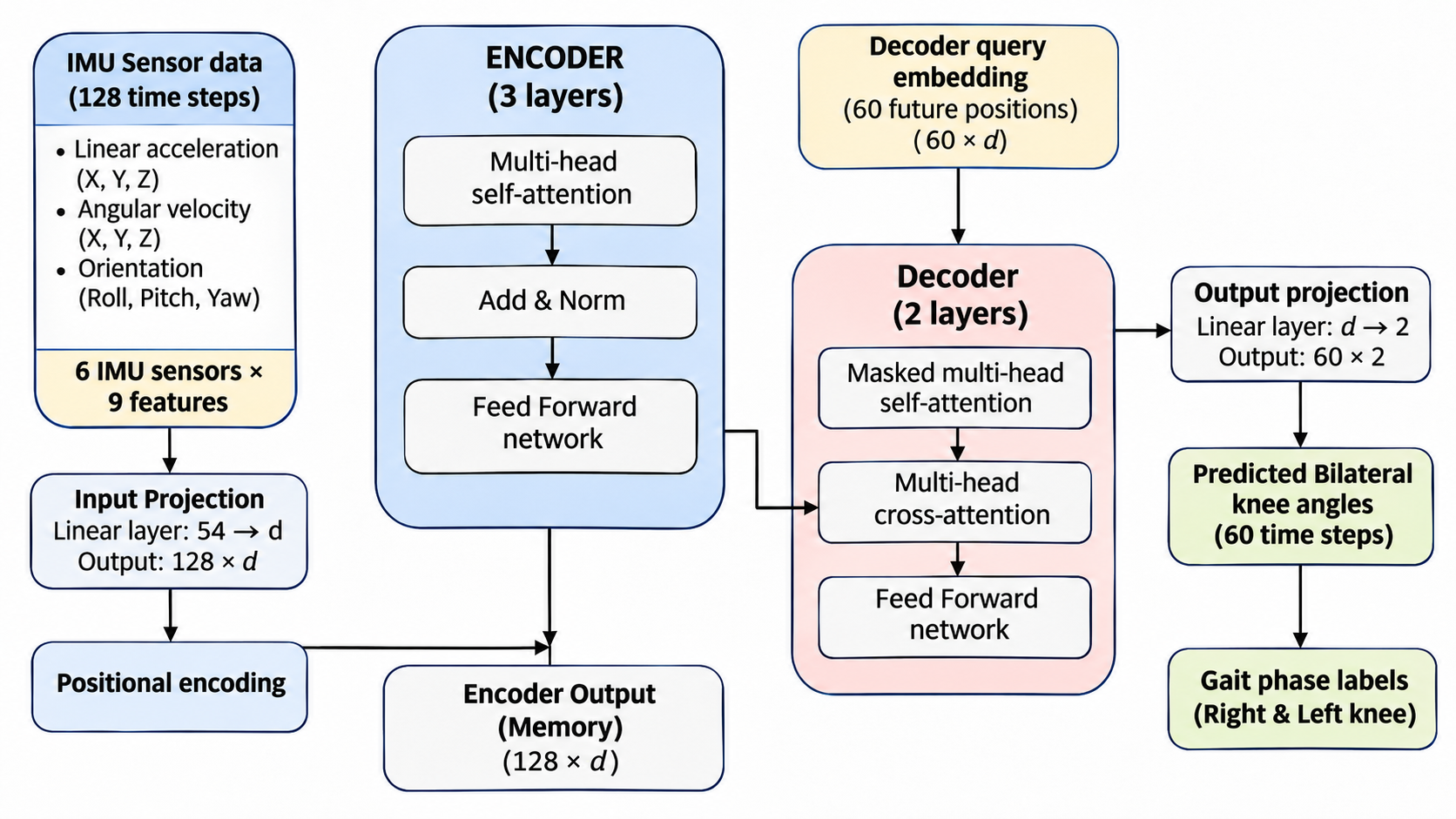}
\caption{Architecture of the proposed Transformer model for bilateral knee joint angle prediction and gait phase estimation}
\label{fig: model_archit}
\end{figure*}

\subsection{Gait Phase Interpretation using Rule-Based Pseudo-Labeling}
\label{subsec:gait_phase_interpretation}

In addition to predicting future bilateral knee joint angles, the proposed framework provides rule-based gait-phase interpretation. As illustrated in Fig.~\ref{fig:methodology_workflow}, gait-phase interpretation constitutes the final stage of the proposed framework. Since the collected dataset did not include manually annotated gait-phase labels, a rule-based pseudo-labeling strategy was developed to assign gait phases from the knee flexion--extension angle trajectories and their temporal characteristics \cite{ref10,ref11}. The procedure was applied independently to both the reference and predicted knee-angle trajectories. The labels derived from the reference knee-angle trajectories were treated as rule-based pseudo-labels for evaluation, rather than as independently annotated ground-truth gait phases.

\begin{figure}[htbp]
\centering
\includegraphics[width=\linewidth]{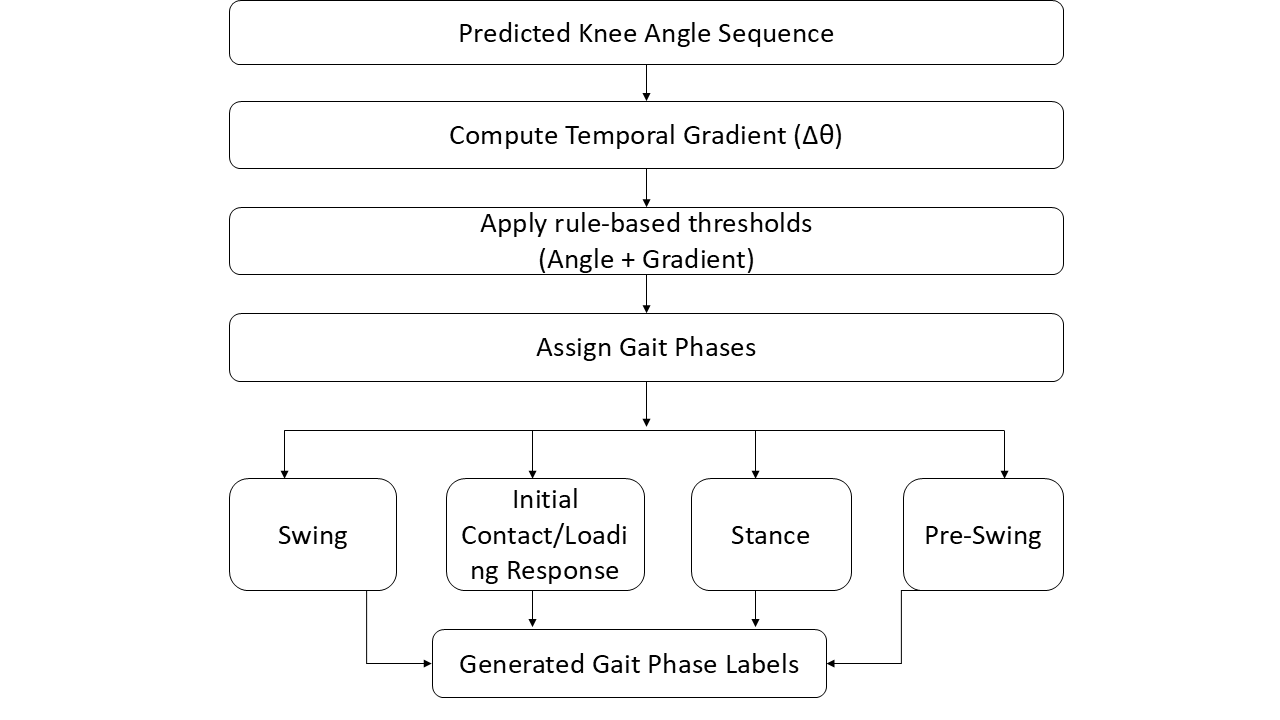}
\caption{Flowchart of the proposed rule-based pseudo-labeling strategy for automatic gait-phase interpretation.}
\label{sec:gait_phase_pseudo_labeling}
\end{figure}

The pseudo-labeling strategy utilizes both the knee flexion--extension angle and its temporal rate of change. Initially, the temporal gradient of the knee-angle sequence is computed to characterize the direction and magnitude of knee movement. Subsequently, the knee angle and its corresponding temporal gradient are jointly analyzed to assign each time step to one of four gait phases according to predefined kinematic rules. The complete pseudo-labeling procedure is illustrated in Fig.~\ref{sec:gait_phase_pseudo_labeling}.

The gait phases are assigned according to the following rules:

\begin{itemize}

\item \textbf{Swing:} Knee flexion angle less than $15^{\circ}$.

\item \textbf{Initial Contact / Loading Response:} Knee flexion angle between $15^{\circ}$ and $30^{\circ}$.

\item \textbf{Stance:} Knee flexion angle between $30^{\circ}$ and $60^{\circ}$ with a temporal gradient $\Delta \theta \leq 1^{\circ}$.

\item \textbf{Pre-Swing:} Knee flexion angle greater than or equal to $60^{\circ}$, or knee flexion angle between $30^{\circ}$ and $60^{\circ}$ with a temporal gradient $\Delta \theta > 1^{\circ}$.

\end{itemize}

The rule-based phase assignment provides a simplified kinematic
interpretation of gait phase based on knee flexion--extension angle
and its temporal gradient. However, knee angle and its temporal
gradient do not uniquely identify a physiological gait phase under
all locomotion conditions. Similar knee-angle values may occur at
different portions of the gait cycle depending on walking speed,
locomotion mode, incline or decline, stair negotiation, and
subject-specific gait strategies. Therefore, the phase labels
generated by the proposed rules should be interpreted as kinematic
pseudo-labels for model interpretation rather than as definitive
physiological or clinically validated gait-phase annotations.

The proposed rule-based pseudo-labeling strategy provides a computationally
efficient and interpretable mechanism for assigning gait phases from
knee-angle trajectories without requiring manually annotated gait
labels. The resulting labels provide a kinematic interpretation of
the predicted knee motion and enable phase-wise analysis of the
prediction results. This allows the proposed framework to combine
future knee-angle prediction with rule-based gait-phase interpretation
while acknowledging the limitations of using knee kinematics alone
for physiological gait-phase identification.

\subsection{Evaluation Metrics}

The performance of the proposed framework was evaluated using Mean
Absolute Error (MAE) and Root Mean Square Error (RMSE) for knee-angle
prediction. MAE and RMSE are defined as

\begin{equation}
MAE =
\frac{1}{N}
\sum_{i=1}^{N}
|y_i-\hat{y}_i|
\end{equation}

\begin{equation}
RMSE =
\sqrt{
\frac{1}{N}
\sum_{i=1}^{N}
(y_i-\hat{y}_i)^2
}
\end{equation}

where $y_i$ and $\hat{y}_i$ denote the reference and predicted
knee-angle values, respectively, and $N$ is the total number of
evaluated samples.

A tolerance-based prediction accuracy was also calculated as the
percentage of predictions within $\pm5^\circ$ of the reference value:

\begin{equation}
A_{5^\circ} =
\frac{1}{N}
\sum_{i=1}^{N}
\mathbb{I}\left(|y_i-\hat{y}_i|\leq5^\circ\right)
\times100
\end{equation}

where $\mathbb{I}$ is 1 when the absolute prediction error is at
most $5^\circ$ and 0 otherwise. MAE and RMSE were used as the primary
regression metrics, while $A_{5^\circ}$ provided a complementary
accuracy measure.

For gait-phase interpretation, pseudo-label agreement accuracy was
calculated as

\begin{equation}
Accuracy_{\mathrm{phase}} =
\frac{N_{\mathrm{correct}}}{N_{\mathrm{total}}}
\times100
\end{equation}

where $N_{\mathrm{correct}}$ is the number of time steps for which
the predicted phase matched the rule-based pseudo-label derived from
the reference knee-angle trajectory. Separate accuracies were
calculated for the right and left knees.

For the confusion matrices, $N_{\mathrm{correct}}$ corresponds to
the sum of the diagonal elements, representing correctly classified
gait-phase samples, while $N_{\mathrm{total}}$ corresponds to the sum
of all elements in the confusion matrix. Thus, the phase accuracy
from each confusion matrix was calculated as

\begin{equation}
Accuracy_{\mathrm{phase}} =
\frac{\sum_i CM_{ii}}
{\sum_i\sum_j CM_{ij}}
\times100
\end{equation}

where $CM_{ij}$ denotes the number of samples belonging to the
reference phase $i$ that were classified as phase $j$. Confusion
matrices were used to visualize phase agreement and disagreement
across the four gait-phase categories.

\section{Results and Discussion}

\subsection{Knee Angle Prediction Performance}

The proposed Transformer-based sequence-to-sequence framework was evaluated for multi-step bilateral knee joint angle prediction using wearable IMU data acquired from the Xsens MVN Awinda\textsuperscript{\textregistered} motion capture system. The model utilized 128 historical time steps as input to predict the subsequent 60 future knee-angle samples, corresponding to a prediction horizon of approximately 0.6~s.

The subject-wise evaluation yielded a Mean Absolute Error (MAE) of
17.54 $\pm$ 6.75$^\circ$ and a Root Mean Square Error (RMSE) of
21.79 $\pm$ 8.23$^\circ$, expressed as mean $\pm$ standard deviation
across the evaluated subjects. The corresponding 95\% confidence
intervals for the mean MAE and RMSE were 14.75--20.33$^\circ$
and 18.38--25.20$^\circ$, respectively. These results characterize
the variability in prediction performance across subjects under
the five-fold subject-level cross-validation protocol. Formal
statistical significance testing was not performed, as the
evaluation focused on prediction accuracy and inter-subject
variability.

The reported MAE and RMSE represent aggregate performance across all
locomotion activities included in the evaluation, namely level-ground
walking, inclined walking, declined walking, stair ascent, and stair
descent. Since knee kinematics may differ across locomotion conditions,
activity-specific performance may provide additional insight into the
generalization of the proposed framework. However, because the
available recordings were not balanced across all locomotion
conditions and subjects, separate activity-wise comparisons were not
used to draw conclusions about the relative difficulty of individual
activities.

The computational performance of the proposed
framework was evaluated using a CUDA-enabled GPU.
As presented in Table~\ref{tab:cpu_gpu},
the average inference latency was 5.82 ms,
with a total testing time of 0.58 s.
The measured inference latency was below the
10 ms sampling interval of the 100 Hz IMU system.
These results indicate the computational feasibility
of the proposed framework for future real-time
gait analysis and wearable assistive applications.
However, real-time performance on an embedded
wearable platform remains to be validated.

\begin{table}[htbp]
\caption{Knee Joint Angle Prediction Performance}
\centering
\begin{tabular}{|l|c|}
\hline
\textbf{Metric} & \textbf{Value} \\
\hline
Sampling Rate & 100 Hz \\
\hline
Input Window Length & 128 Time Steps \\
\hline
Future Prediction Length & 60 Time Steps \\
\hline
Future Start Delay & 20 ms \\
\hline
Training Time & 172.7 s \\
\hline
Testing Time & 0.58 s \\
\hline
Pooled Sample-Level MAE & 1.95$^\circ$ \\
\hline
Pooled Sample-Level RMSE & 2.65$^\circ$ \\
\hline
\end{tabular}
\label{tab:knee_prediction_results}
\end{table}

Table~\ref{tab:knee_prediction_results} presents the
prediction performance obtained from the original
experimental evaluation. The pooled sample-level
MAE and RMSE were 1.95$^\circ$ and 2.65$^\circ$,
respectively. These results are reported separately
from the five-fold subject-level cross-validation
results presented in Section~4.1, which evaluate
prediction performance on previously unseen subjects.
The two sets of results reflect different evaluation
protocols and should not be directly compared.

\begin{figure*}[t]
\centering
\includegraphics[width=0.80\textwidth]{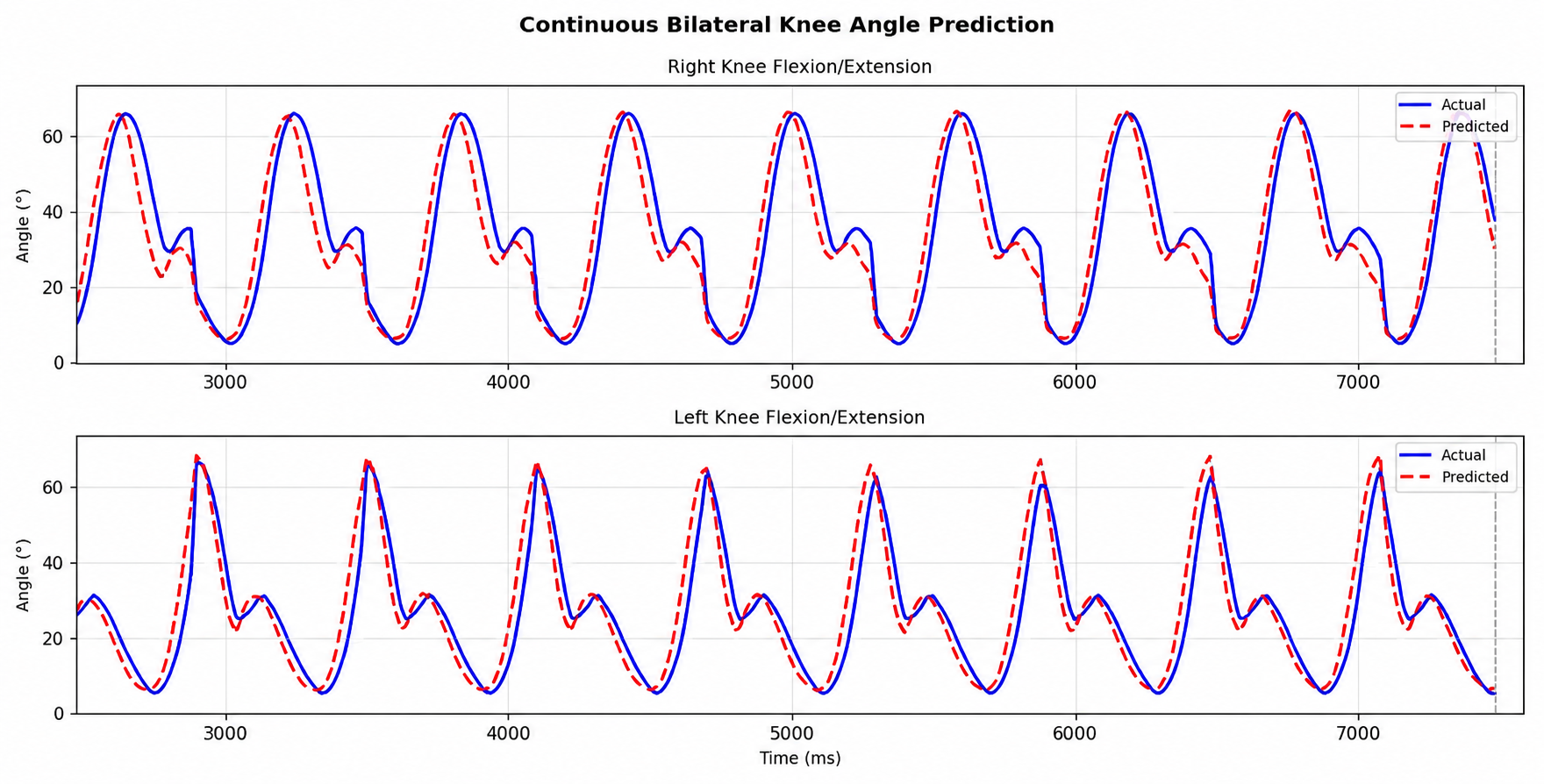}
\caption{Actual and predicted bilateral knee joint angles obtained using the proposed Transformer model.}
\label{fig:transformer_prediction}
\end{figure*}

Figure~\ref{fig:transformer_prediction} illustrates the predicted and
reference bilateral knee-angle trajectories for a representative
prediction window from the test set. The predicted trajectories
generally follow the temporal evolution of the reference knee motion,
while local deviations are visible at certain time points.

The trajectory shown in Fig.~\ref{fig:transformer_prediction}
represents a representative prediction window from the test set and
is therefore not intended to depict the overall prediction error.
The reported MAE and RMSE in Table~\ref{tab:knee_prediction_results}
are calculated over all evaluated prediction samples in the test set.
Consequently, local deviations visible in an individual prediction
window may be larger than the corresponding overall average error.

\subsection{Computational Performance Comparison (CPU vs GPU)}

To evaluate the computational efficiency of the proposed Transformer-based framework, experiments were conducted on both a conventional CPU and a CUDA-enabled GPU using the same dataset, network architecture, and training configuration. This comparison was performed to investigate the effect of hardware acceleration on model training and inference while ensuring that the predictive performance remained unchanged.

The quantitative comparison is presented in Table~\ref{tab:cpu_gpu}. GPU execution substantially reduced the computational cost associated with both model training and inference. The training time decreased from 5866.33~s on the CPU to 172.71~s on the GPU, corresponding to an approximately 33.97$\times$ speed improvement. Similarly, the testing time was reduced from 32.78~s to 0.58~s, while the average inference latency decreased from 23.81~ms to 5.82~ms.

\begin{table}[H]
\caption{CPU and CUDA-Enabled GPU Performance Comparison}
\label{tab:cpu_gpu}
\centering
\begin{tabular}{|l|c|c|c|}
\hline
\textbf{Metric} & \textbf{CPU} & \textbf{GPU (CUDA)} & \textbf{Improvement} \\
\hline
Training Time (s) & 5866.33 & 172.71 & 33.97$\times$ Faster \\
\hline
Testing Time (s) & 32.78 & 0.58 & 56.52$\times$ Faster \\
\hline
Average Latency (ms) & 23.81 & 5.82 & 4.09$\times$ Lower \\
\hline
\end{tabular}
\end{table}

The results demonstrate that CUDA-based parallel computation significantly accelerates the execution of the proposed framework while preserving prediction accuracy. Such computational improvements are particularly important for future real-time gait analysis and wearable assistive systems, where rapid model inference is essential for timely decision making.

\subsection{Comparison with Sequential Baseline Models}

To assess the prediction capability of the proposed
Transformer-based framework, its performance was
compared with three sequential baseline models:
Long Short-Term Memory (LSTM), Gated Recurrent
Unit (GRU), and Temporal Convolutional Network
(TCN)\cite{ref5}. The reported results provide a comparison
of prediction errors under the respective experimental
configurations. Since the models differ in their
input-window lengths and forecasting strategies,
the comparison is intended to provide a performance
reference rather than establish architectural
superiority under identical conditions..

The LSTM used a 30-sample (0.3~s) historical window and generated the
60-sample future trajectory through recursive one-step prediction.
The GRU and TCN used the same 30-sample input window and directly
predicted 60 future samples in a single forward pass. In contrast, the
proposed Transformer used 128 previous samples (1.28~s) as input and
directly predicted 60 future samples (0.6~s). Thus, the comparison
considers different forecasting strategies rather than a strictly
one-to-one architectural comparison.

\begin{figure*}[!t]
\centering
\includegraphics[width=0.80\textwidth]{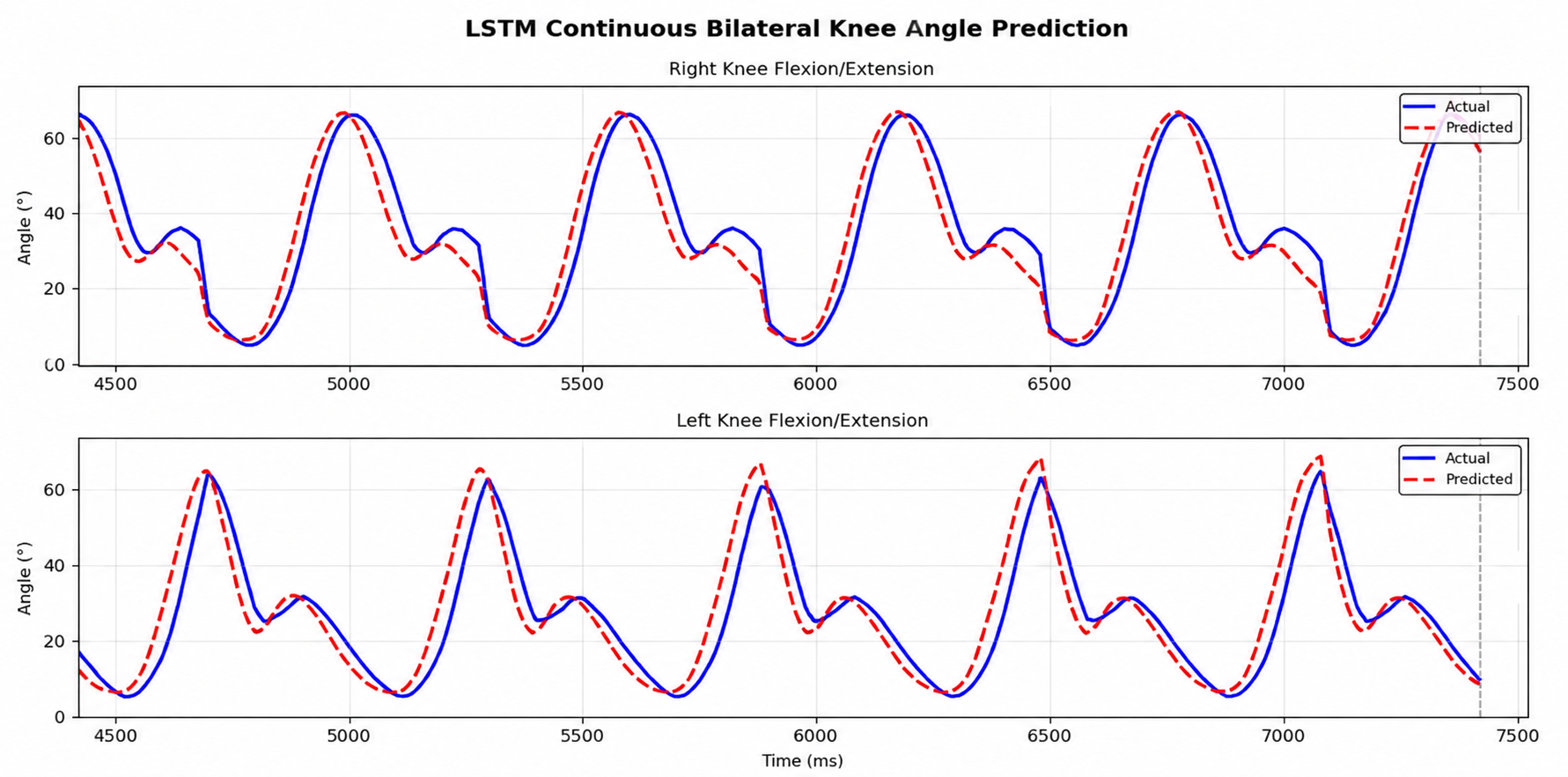}
\caption{Actual and predicted bilateral knee joint angles obtained using
the LSTM model with recursive 60-step forecasting.}
\label{fig:lstm_prediction}
\end{figure*}

Table~\ref{tab:model_comparison} summarizes the input configuration,
forecasting strategy, prediction horizon, and quantitative prediction
performance of the evaluated models.

\begin{table}[htbp]
\caption{Comparison of LSTM, GRU, TCN, and Transformer-Based Knee Angle Prediction Models}
\label{tab:model_comparison}
\centering
\renewcommand{\arraystretch}{1.15}
\resizebox{\columnwidth}{!}{%
\begin{tabular}{|p{2.7cm}|p{3.2cm}|c|c|c|}
\hline
\textbf{Model} & \textbf{Prediction Strategy} & \textbf{Input Window} & \textbf{MAE ($^\circ$)} &
\textbf{RMSE ($^\circ$)}\\
\hline
LSTM &
60-step Recursive &
30 samples (0.3 s) &
5.70 &
7.90 \\
\hline

GRU &
60-step Direct &
30 samples (0.3 s) &
11.11 &
17.32 \\
\hline

TCN &
60-step Direct &
30 samples (0.3 s) &
10.72 &
16.54 \\
\hline

Transformer &
60-step Direct &
128 samples (1.28 s) &
1.95 &
2.65 \\
\hline
\end{tabular}
}
\end{table}

As shown in Table~\ref{tab:model_comparison}, the Transformer achieved
the lowest prediction error, with an MAE of 1.95$^\circ$ and an RMSE of
2.65$^\circ$. The LSTM achieved an MAE of 5.70$^\circ$ and an RMSE of
7.90$^\circ$, while the TCN and GRU achieved MAEs of 10.72$^\circ$ and
11.11$^\circ$, respectively. The results indicate lower prediction
errors for the Transformer under the evaluated configurations;
however, they should not be interpreted as a direct claim of
architectural superiority because the models use different input
windows and forecasting strategies.

\subsection{Gait Phase Prediction Performance}

In addition to future bilateral knee-angle prediction, the proposed framework was evaluated for automatic gait-phase interpretation. Since manually annotated gait-phase labels were not available in the dataset, the rule-based pseudo-labeling strategy described in Section~\ref{subsec:gait_phase_interpretation} was employed to generate gait-phase labels directly from the predicted knee-angle trajectories.

The gait cycle was divided into four biomechanically meaningful phases, namely Swing, Initial Contact/Loading Response, Stance, and Pre-Swing. The generated gait-phase labels were compared with the corresponding reference labels to evaluate the effectiveness of the proposed gait interpretation module.

\begin{figure*}[!t]
\centering

\begin{subfigure}{0.45\textwidth}
    \centering
    \includegraphics[width=\linewidth]{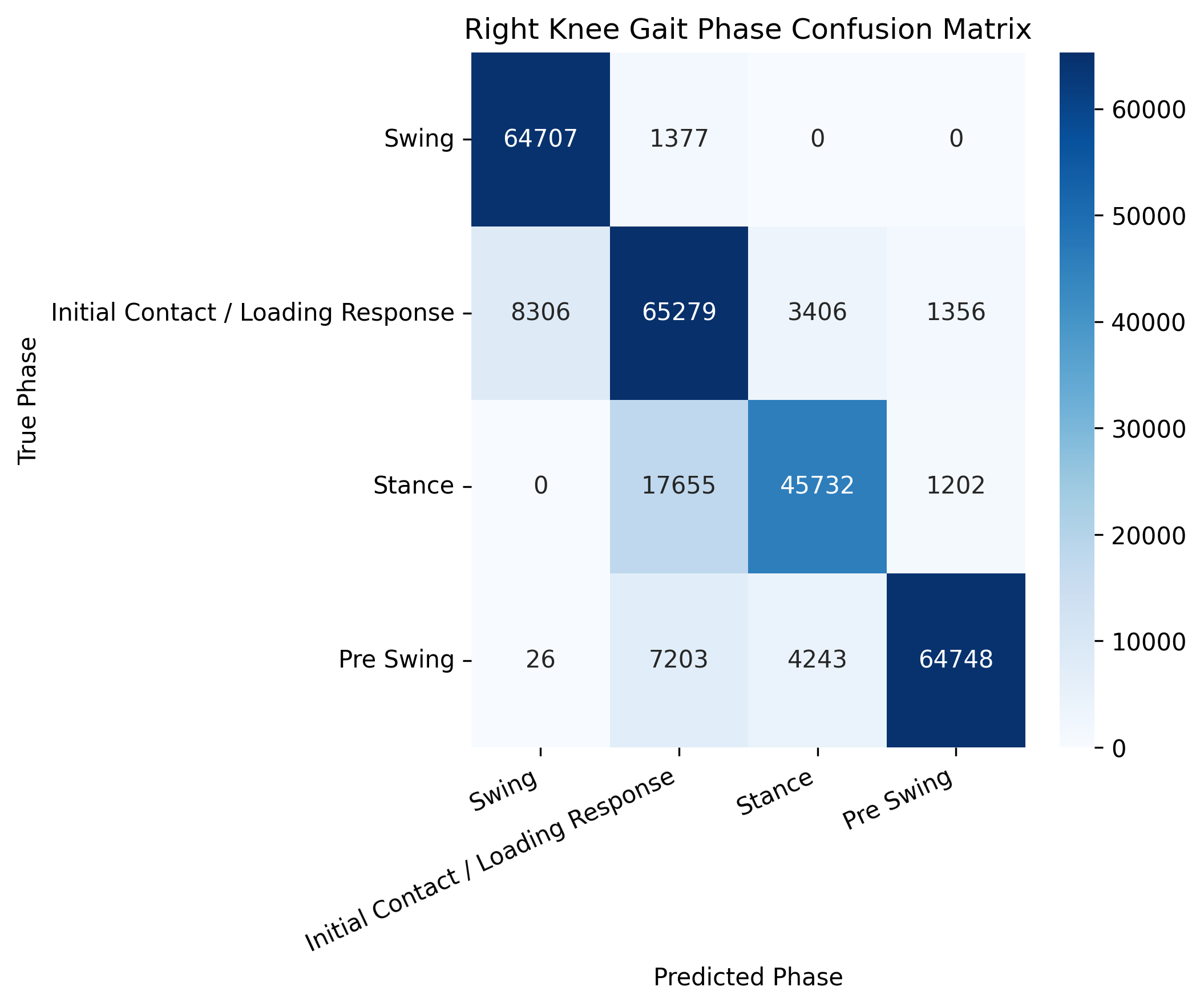}
    \caption{Right Knee}
\end{subfigure}
\hfill
\begin{subfigure}{0.45\textwidth}
    \centering
    \includegraphics[width=\linewidth]{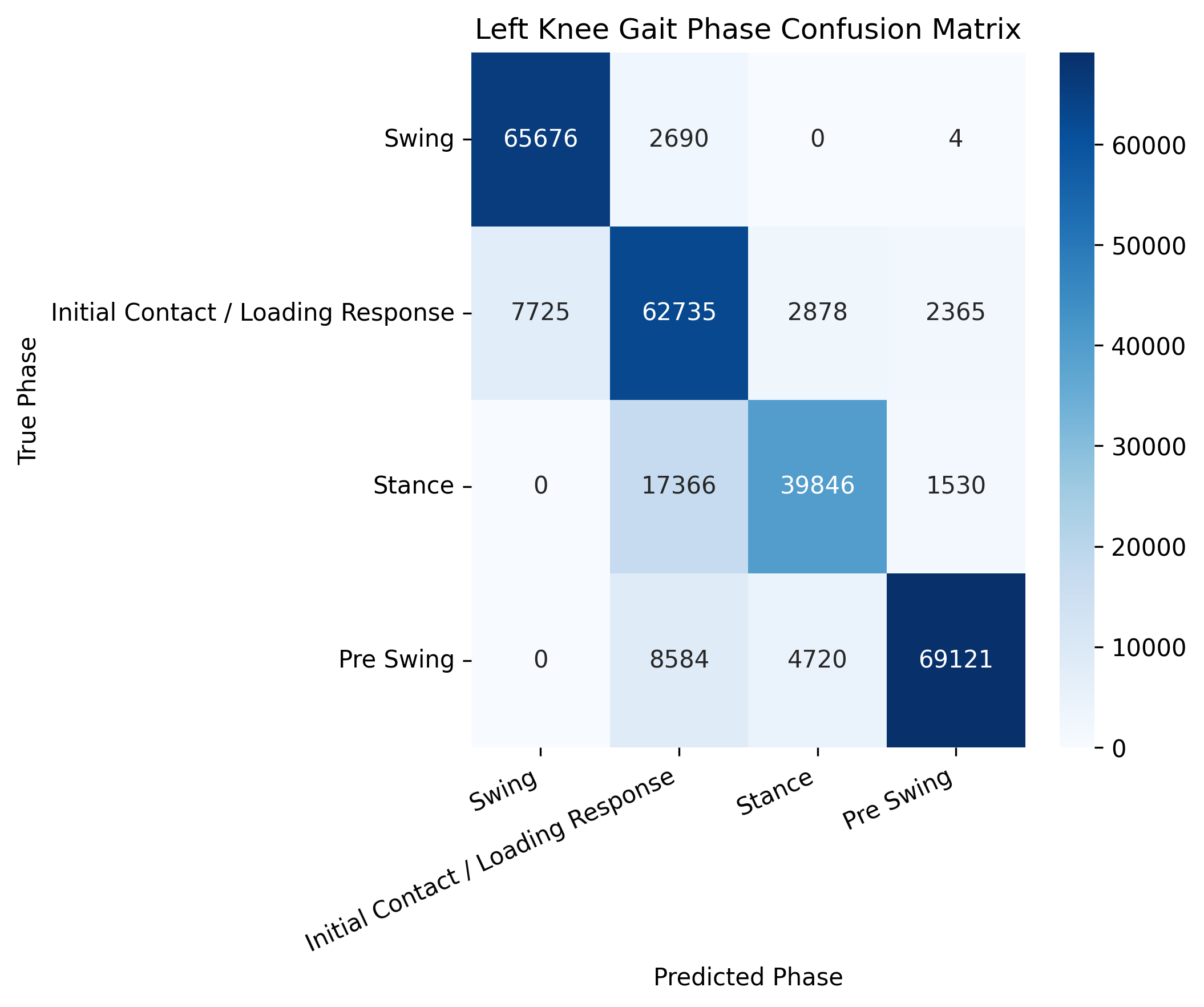}
    \caption{Left Knee}
\end{subfigure}

\caption{Confusion matrices for bilateral knee gait phase classification using the proposed Transformer model}
\label{fig:confusion_matrices}
\end{figure*}

Table~\ref{tab:gait_accuracy} summarizes the
gait-phase interpretation performance for both knee joints.
The proposed framework achieved an accuracy of 84.41\%
for the right knee and 83.22\% for the left knee, with an
average accuracy of 83.82\%. These values were calculated
from the confusion matrices presented in Fig.~\ref{fig:confusion_matrices}.
The results indicate agreement between the gait-phase
pseudo-labels derived from the predicted and reference
knee-angle trajectories.

\begin{table}[htbp]
\caption{Gait Phase Interpretation Performance}
\label{tab:gait_accuracy}
\centering
\renewcommand{\arraystretch}{1.2}
\begin{tabular}{|l|c|}
\hline
\textbf{Metric} & \textbf{Value} \\
\hline
Number of Gait Phases & 4 \\
\hline
Pseudo-Labeling Method & Rule-Based \\
\hline
Right Knee Accuracy & 84.41\% \\
\hline
Left Knee Accuracy & 83.22\% \\
\hline
Average Accuracy & 83.82\% \\
\hline
\end{tabular}
\end{table}

Figure~\ref{fig:confusion_matrices} presents the confusion matrices for bilateral gait-phase interpretation. Most samples were correctly assigned to their respective gait phases, as indicated by the strong concentration of values along the main diagonal. The Swing and Pre-Swing phases exhibited the highest classification performance, whereas a limited number of misclassifications occurred near phase-transition regions, particularly between Swing and Initial Contact/Loading Response. Such errors are expected because adjacent gait phases share similar biomechanical characteristics during transitional movements.

A small number of misclassifications occurred near gait-phase transition regions, especially between Swing and Initial Contact/Loading Response. Such errors are expected because adjacent gait phases often share similar movement characteristics during transitions \cite{ref10,ref11}. Despite these challenges, the overall classification results indicate that the proposed approach can reliably interpret gait behaviour from predicted knee motion.

\begin{figure*}[!t]
\centering
\includegraphics[width=0.80\textwidth]{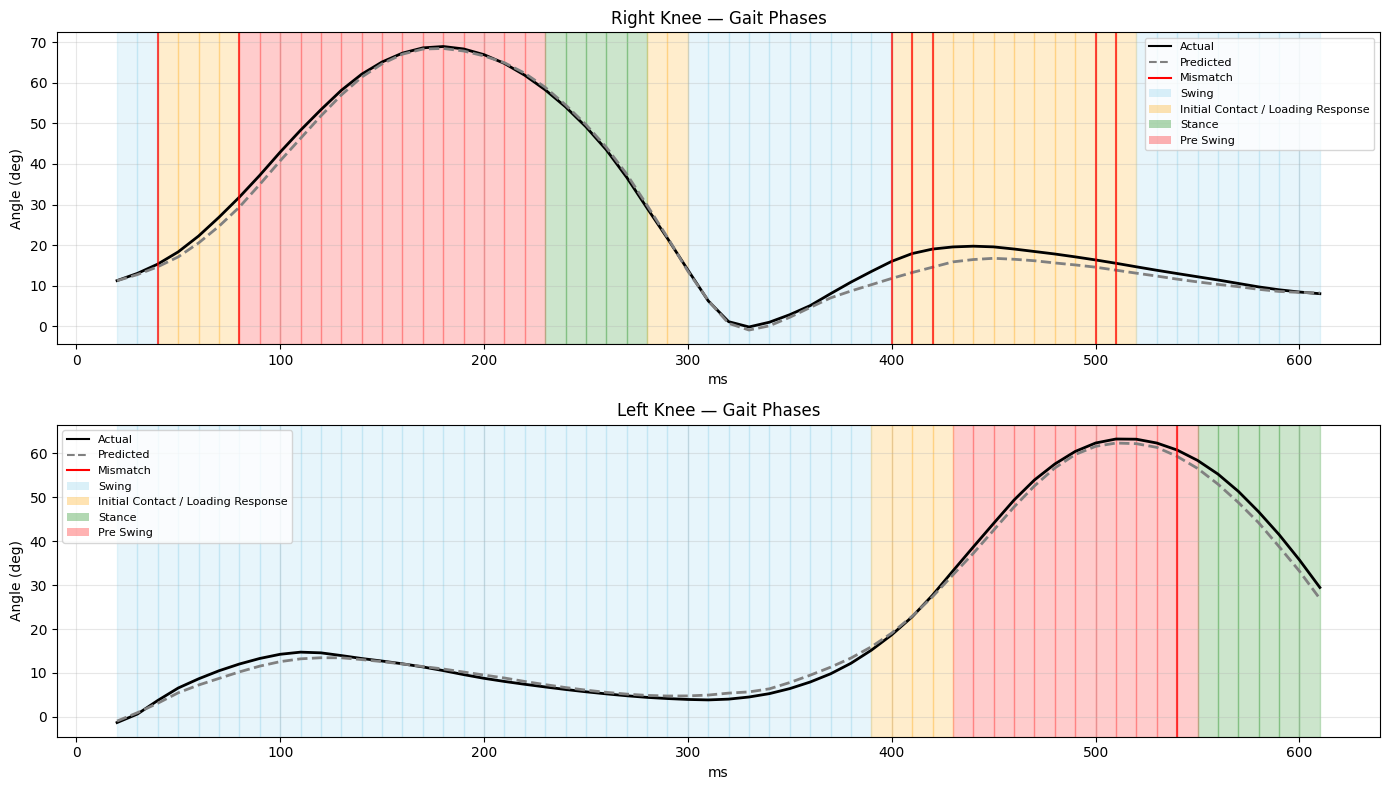}
\caption{Bilateral knee joint angle trajectories with the corresponding predicted gait phases}
\label{fig:gait_phase_visualization}
\end{figure*}

Figure~\ref{fig:gait_phase_visualization} illustrates the predicted gait phases superimposed on the corresponding bilateral knee-angle trajectories. The close correspondence between the gait-phase labels and the temporal evolution of the knee-angle profiles demonstrates that the proposed framework successfully combines accurate future motion prediction with meaningful gait interpretation within a unified architecture.

\subsection{Discussion}

The experimental results demonstrate that the proposed Transformer-based sequence-to-sequence framework effectively learned the temporal characteristics of bilateral lower-limb motion and accurately predicted future knee-angle trajectories from wearable IMU data. By employing a multi-head self-attention mechanism, the model was able to capture both local and long-range temporal dependencies without relying on sequential hidden-state propagation, making it well suited for long-horizon gait prediction.

Compared with the baseline LSTM model, the proposed Transformer employs a sequence-to-sequence architecture to generate 60 future knee-angle samples in a single forward pass. In contrast, the LSTM baseline uses recursive forecasting to generate the corresponding prediction horizon. As reported in Table VI, the Transformer achieved lower prediction errors in the original experimental evaluation. However, differences in forecasting strategies and experimental configurations limit direct architectural comparisons. The additional subject-wise cross-validation results further highlight the challenges of generalizing knee-angle prediction to previously unseen individuals.

The proposed rule-based pseudo-labeling strategy enabled gait-phase interpretation from predicted bilateral knee-angle trajectories without requiring manually annotated gait-phase labels. The agreement accuracies of 84.41\% and 83.22\% for the right and left knees, respectively, indicate that the predicted trajectories preserve temporal characteristics relevant to gait-phase interpretation. However, these results represent agreement with kinematically derived pseudo-labels and should not be interpreted as independently validated clinical gait-phase recognition performance.

From a clinical perspective, accurate prediction of future knee joint motion has the potential to assist clinicians in monitoring rehabilitation progress, evaluating gait abnormalities, and assessing functional recovery. Furthermore, predicting future motion may help compensate for sensing, computation, and actuation delays in wearable robotic systems, thereby enabling smoother and more responsive assistance during walking.

Beyond clinical rehabilitation, the proposed framework has potential applications in wearable assistive devices, lower-limb exoskeletons, intelligent prosthetic systems, human motion analysis, and real-time gait monitoring. The ability to simultaneously perform long-horizon knee-angle prediction and gait-phase interpretation within a unified framework demonstrates the potential of Transformer-based models for future intelligent wearable healthcare technologies.

\subsection{Limitations}

Despite the promising results, the proposed framework has several limitations that should be considered.

\begin{itemize}

\item The experimental evaluation was conducted using data collected exclusively from healthy participants. Therefore, the generalizability of the proposed framework to individuals with gait impairments or neurological disorders remains to be investigated.

\item The dataset comprised a limited range of locomotion activities, including level-ground walking, stair ascent, stair descent, inclined walking, and declined walking. Additional locomotion scenarios, such as obstacle negotiation, turning, running, and irregular terrain, were not considered.

\item The proposed framework utilized only wearable IMU sensor data. Incorporating additional sensing modalities, such as surface EMG, plantar pressure sensors, or force sensors, may further improve motion prediction and gait-phase interpretation.

\item The reference knee-angle trajectories were obtained from the Xsens MVN Awinda\textsuperscript{\textregistered} sensor-fusion system. Therefore, the reported prediction errors represent agreement with Xsens-derived knee kinematics rather than validation against an independent gold-standard motion-capture system. Independent validation using marker-based motion capture or another established reference measurement would provide stronger evidence of the absolute accuracy of the predicted knee angles.

\item The proposed framework was evaluated under offline experimental conditions and has not yet been validated in a real-time wearable rehabilitation or assistive robotic system.

\item The five-fold subject-wise cross-validation produced higher prediction errors (MAE: 17.54 ± 6.75°; RMSE: 21.79 ± 8.23°) than the original experimental evaluation. This indicates that prediction performance is sensitive to inter-subject variability and highlights the need for improved generalization to previously unseen individuals.

\item Gait-phase performance was evaluated using rule-based pseudo-labels derived from knee-angle trajectories rather than independently annotated physiological gait events. Therefore, the reported agreement accuracies do not establish clinical gait-phase recognition accuracy. Further validation against independently annotated gait events is required.

\end{itemize}

\subsection{Future Work}

Future research will focus on extending the proposed framework in several directions.

\begin{itemize}

\item Integrating wearable IMU data with additional sensing modalities, such as surface EMG and plantar pressure sensors, to further improve knee-angle prediction and gait-phase interpretation.

\item Evaluating the proposed framework using larger and more diverse datasets involving different age groups, walking speeds, and locomotion activities to improve model generalizability.

\item Validating the proposed framework using data collected from individuals with gait impairments and neurological disorders to assess its clinical applicability.

\item Implementing and evaluating the proposed framework in a real-time wearable system for continuous gait monitoring and intelligent rehabilitation assistance.

\item Investigating uncertainty-aware motion prediction and adaptive gait analysis to improve prediction reliability under complex locomotion conditions.

\item Extending the proposed framework for applications in lower-limb exoskeletons, intelligent prosthetic systems, and wearable rehabilitation technologies.

\item Investigating subject-adaptive learning, transfer learning, and domain adaptation strategies to improve prediction accuracy and generalization across previously unseen individuals, particularly in view of the performance variability observed during subject-wise cross-validation.

\end{itemize}

\section{Conclusion}

\

This study presented a Transformer-based sequence-to-sequence framework for multi-step bilateral knee joint angle prediction and rule-based gait-phase interpretation using wearable IMU sensor data acquired from 25 healthy participants. The proposed architecture utilized 128 historical time steps to predict 60 future samples of right and left knee flexion–extension angles, incorporating a 20 ms prediction gap.

The original experimental evaluation achieved a mean absolute error (MAE) of 1.95° and a root mean square error (RMSE) of 2.65°. In contrast, five-fold subject-wise cross-validation yielded an MAE of 17.54 ± 6.75° and an RMSE of 21.79 ± 8.23°, highlighting the challenges of generalization to previously unseen subjects. These results were obtained using different evaluation protocols and should not be interpreted as directly comparable estimates of model performance.

Rule-based gait-phase interpretation achieved pseudo-label agreement accuracies of 84.41\% and 83.22\% for the right and left knees, respectively, with an average agreement of 83.82\%. These findings indicate that predicted knee-angle trajectories retain temporal information relevant to gait-phase interpretation, although independent validation using annotated physiological gait events remains necessary.

Overall, the proposed framework demonstrates the feasibility of integrating multi-step bilateral knee-angle prediction and gait-phase interpretation within a unified architecture. Future investigations will focus on improving subject-independent generalization, validating the framework using clinical gait datasets, and evaluating real-time deployment in wearable rehabilitation and lower-limb assistive robotic systems.

\section{Acknowledgment}

The authors express their sincere gratitude to the Director, Defence Bioengineering and Electromedical Laboratory (DEBEL), Defence Research and Development Organisation (DRDO), Bengaluru, for providing the necessary research facilities and institutional support to carry out this study. The authors also acknowledge the valuable guidance and technical assistance provided by the scientists, researchers, engineers, and technical staff of DEBEL, DRDO, during the course of this research.

\end{document}